\documentclass[10pt,twocolumn,letterpaper]{article}

\usepackage{multirow}
\usepackage{float}
\usepackage{placeins}
\usepackage[pagenumbers]{wacv}

\usepackage{graphicx}
\usepackage{amsmath}
\usepackage{amssymb}
\usepackage{booktabs}
\usepackage[accsupp]{axessibility}

\usepackage[pagebackref,breaklinks,colorlinks]{hyperref}

\usepackage[capitalize]{cleveref}
\crefname{section}{Sec.}{Secs.}
\Crefname{section}{Section}{Sections}
\Crefname{table}{Table}{Tables}
\crefname{table}{Tab.}{Tabs.}

\def\wacvPaperID{2687} 
\def\confName{WACV}
\def\confYear{2025}

\begin{document}

\title{Multispecies Animal Re-ID Using a Large Community-Curated Dataset}

\author{
  {\small \textbf{Lasha Otarashvili}\thanks{\texttt{lasha@conservationxlabs.org}}} \\
  {\small Conservation X Labs}
  \and
  {\small \textbf{Tamilselvan Subramanian}} \\
  {\small Conservation X Labs}
  \and
  {\small \textbf{Jason Holmberg}} \\
  {\small Conservation X Labs}
  \and
  {\small \textbf{J.J. Levenson}} \\
  {\small Bureau of Ocean Energy Management: Washington, DC, US}
  \and
  {\small \textbf{Charles V.\ Stewart}} \\
  {\small Rensselaer Polytechnic Institute}
}

\maketitle

\begin{abstract}
Recent work has established the ecological importance of developing algorithms for identifying animals individually from images. Typically, a separate algorithm is trained for each species, a natural step but one that creates significant barriers to wide-spread use: (1) each effort is expensive, requiring data collection, data curation, and model training, deployment, and maintenance, (2) there is little training data for many species, and (3) commonalities in appearance across species are not exploited.
We propose an alternative approach focused on training multi-species individual identification (re-id) models. We construct a dataset that includes 49 species, 37K individual animals, and 225K images, using this data to train a single embedding network for all species. Our model employs an EfficientNetV2 backbone and a sub-center ArcFace loss function with dynamic margins. We evaluate the performance of this multi-species model in several ways. Most notably, we demonstrate that it consistently outperforms models trained separately on each species, achieving an average gain of 12.5\% in top-1 accuracy. Furthermore, the model demonstrates strong zero-shot performance and fine-tuning capabilities for new species with limited training data, enabling effective curation of new species through both incremental addition of data to the training set and fine-tuning without the original data. Additionally, our model surpasses the recent MegaDescriptor on unseen species, averaging an 19.2\% top-1 improvement per species and showing gains across all 33 species tested. The fully-featured code repository is publicly available on GitHub, and the feature extractor model can be accessed on HuggingFace for seamless integration with wildlife re-identification pipelines. The model is already in production use for 60+ species in a large-scale wildlife monitoring system.\cite{wildme_platforms},
\end{abstract}

\section{Introduction}

Individual animal identification has significant challenges that are not typically encountered in related problems such as human face recognition \cite{Deng_2022, wang2022surveyfacerecognition} and human re-id \cite{ye2021deeplearningpersonreidentification}:
\begin{itemize}
\item Ground truth training datasets tend to be small. This is because individual animal identity is difficult and time-consuming to curate: it generally relies on humans searching for matching images, an effort that grows quadratically with the size of the dataset.
\item Some species have large volumes of training data while others have relatively little. This is partially caused by difficult \textit{in situ} imaging conditions that make it hard to obtain good quality photos.  It is also a result of the aforementioned challenges of manual curation, where the resources necessary to build large training sets are available for only the most charismatic species.
\item Standard practice is to develop and train animal re-id algorithms on a species-by-species basis, sometimes with per-species algorithmic features. This practice is reasonable in research settings and when just a few species are considered, but it is increasing impractical when supporting dozens of species, each with frequent, ongoing additions of new images and individuals. 
\end{itemize}

Our goal in this paper is to address these issues through training and testing a single model for identifying individual animals from many different species.
We go beyond the ground breaking work of \cite{megadescriptor} and explore a dataset that is expert-curated from ongoing use of animal re-id in scientific and conservation practice.
Our study focuses on two broad issues:
\begin{itemize}
\item How does the overall performance of models trained on multiple species compare to those trained on individual species? When performance is equivalent then individual species models are unnecessary, and when performance is better on the multiple species model we also see a performance advantage to combined training on multiple species.
\item How well do multispecies models perform when applied to species unseen during training? This is crucial to addressing the shortage of annotated training data for many species.  Ideally, a model would work as well as if training data were available, but practically if a multi-species model can significantly reduce the amount of training data needed obtain a given level of performance for a new species it is an important development. 
\end{itemize}
As a result of our study, we make the following main contributions:
\begin{enumerate}
    \item We train a multispecies animal re-id model MiewID on a community-curated dataset involving 49 species from 59 datasets, 37,138 individual animals and 225,374 total annotations. We publicly release the code as a \href{https://github.com/WildMeOrg/wbia-plugin-miew-id}{GitHub repository} and \href{https://huggingface.co/conservationxlabs/miewid-msv2}{model weights on Huggingface}. We show that our model consistently outperforms analogous single-species models, exhibiting an overall performance gain of 12.5\% top-1 accuracy and per-species gains ranging from a minimum of -0.2\% (negative for 1 species) to a maximum of 77.3\%.
    \item We demonstrate the effectiveness of our model on previously species unseen during training, and show that it substantially outperforms MegaDescriptor \cite{megadescriptor} on all 33 tested species, with an average top-1 improvement of 19.2\%.
    \item We show the potential of our model for learning to effectively identify species that 
    have only have a small set of data available during training. This is seen both when the new data are incorporated into the overall training set and when fine-tuning the published model.
\end{enumerate}

\section{Background}

Historically, a wide range of methods have been used for animal re-identification, ranging from manual feature extraction to more recent advances in deep learning. In recent years, global descriptor-based methods leveraging deep learning \cite{dlamini2020automated, piev2, megadescriptor, wildfusion, towards, dogfacenet} have gained prominence over traditional manual feature-based approaches \cite{curvrank, norpppa, finfindr} and local feature algorithms \cite{hotspotter, norpppa, lightglue, superglue}. These deep learning models are becoming increasingly popular due to their ability to produce more robust, generalizable descriptors compared to hand-crafted or localized feature techniques.
A major advantage of deep learning methods is their scalability and capability to streamline the re-identification pipeline. While they require larger datasets and greater computational resources to train, they minimize the need for extensive preprocessing and can run end-to-end, making them highly efficient for large-scale animal ID tasks. In this work, we focus on end-to-end deep learning models due to their ability to handle complex identification tasks across multiple species with minimal manual intervention, thus improving scalability and applicability to real-world scenarios.
The release of several large animal re-identification datasets has accelerated research in this field \cite{Cheeseman2022-vd, atrw_dataset, freytag2016chimpanzee, 10.1007/978-3-030-29894-4_34, primface_dataset, witham2019macaque, s22197602, Adam_2024_WACV, dlamini2020automated, lila_science, botswana2022, belugaid}. These datasets cover a broad spectrum of species, providing the necessary data to train more effective and generalizable models. Recent studies have explored multi-species re-identification, which focuses on applying individual identification methods across diverse sets of species \cite{Cheeseman2022-vd, megadescriptor, towards}.
Our work reported here goes beyond these methods in several important ways, including the quality and diversity of species used in training and the extensive experiments in zero-shot and few-shot fine-tuning.

%
%
%

\section{Data}

We construct our multi-species dataset from a mixture of 1) publicly available datasets\cite{Cheeseman2022-vd,atrw_dataset,freytag2016chimpanzee,10.1007/978-3-030-29894-4_34,primface_dataset,witham2019macaque,s22197602,Adam_2024_WACV,dlamini2020automated} and 2) ecological data and images contributed directly and privately to the Wild Me Lab's self-hosted Wildbook platforms\cite{wildme_platforms}, which support dynamic and secure data curation (to the individual animal level) by corresponding species experts. Inside Wildbook, users employ manual photo ID (i.e., ``by eye'') and earlier re-identification computer vision algorithms 
\cite{hotspotter,Weideman2017IntegralCR,Moskvyak2019RobustRO} to rank-order potential individual matches and then make curation decisions about individual animal identities across sightings directly in the software. Our work herein is in direct assistance to the Wildbook user community (2000+ users), which we transparently support via the Wild Me community platform (\url{https://community.wildme.org/}). Contributed data is used with permission to correspondingly improve individual matching capabilities in the Wildbook software platform. Where expressly permitted by data owners using Wildbook, we contribute these unique, species-specific re-ID datasets\cite{beluga_id_2022,hyena_id_2022,leopard_id_2022,whale_shark_id,parham2017animal} back for public access on LILA BC (\url{https://lila.science/}).
Statistics on the individual species are summarized in Table~\ref{table:data_summary}.

Overall, there is an average of 6.25 annotations per individual and a median of 3.  Clearly, this is a heavy-tailed distribution, with five species having more than 10,000 annotations, 11 having fewer than 1,000, 9 fewer than 500, and 6 fewer than 250.

\begin{table}[htbp]
\caption{Dataset summary, including name, source, number of annotations and number of individuals. 
Different image sets for a single species are combined when forming the training and test sets.}
\label{table:data_summary}
\centering
\resizebox{\columnwidth}{!}{
\begin{tabular}{l l l l}
\toprule
Dataset & Source & Annotations & Individuals \\ \midrule
amur\_tiger & contributed& 1,015 & 103 \\
beluga\_whale & contributed& 4,055 & 681 \\
blue\_whale & Happywhale & 4,830 & 2348 \\
bottlenose\_dolphin & contributed & 22,691 & 4,130 \\
brydes\_whale & Happywhale & 154 & 44 \\
capuchin & Lomas Capuchin & 2,618 & 35 \\
cheetah & contributed & 3,005 & 403 \\
chimpanzee & contributed & 203 & 11 \\
chimpanzee\_ctai & C-Tai & 4,662 & 71 \\
chimpanzee\_czoo & C-Zoo & 2,109 & 24 \\
commersons\_dolphin & Happywhale & 90 & 70 \\
cuviers\_beaked\_whale & Happywhale & 341 & 183 \\
dog & DogFaceNet & 8363 & 1,392 \\
dusky\_dolphin & Happywhale & 3,139 & 2,731 \\
eurasianlynx & contributed& 4,716 & 1,642 \\
fin\_whale & Happywhale & 1,324 & 466 \\
frasiers\_dolphin & Happywhale & 14 & 13 \\
giraffe & contributed& 3,724 & 1,067 \\
horse\_wild+face & contributed& 293 & 115 \\
horse\_wild\_tunisian+face & contributed& 1,403 & 47 \\
hyena & contributed& 2,948 & 498 \\
hyperoodon\_ampullatus & contributed& 30,063 & 1,196 \\
jaguar & contributed& 1,876 & 257 \\
japanese\_monkey & PrimFace & 297 & 18 \\
killer\_whale & Happywhale & 2,455 & 472 \\
leopard & contributed& 6,052 & 848 \\
lion & contributed& 6,374 & 397 \\
long\_finned\_pilot\_whale & Happywhale & 238 & 131 \\
lynx\_pardinus & contributed& 2,573 & 752 \\
macaque\_face & MacaqueFaces & 6,280 & 34 \\
melon\_headed\_whale & Happywhale & 1,689 & 1,323 \\
nyala & contributed& 1,942 & 237 \\
pantropic\_spotted\_dolphin & Happywhale & 145 & 44 \\
pygmy\_killer\_whale & Happywhale & 76 & 24 \\
rhesus\_monkey & PrimFace & 763 & 38 \\
rough\_toothed\_dolphin & Happywhale & 60 & 46 \\
seal & SealID & 2,080 & 57 \\
sei\_whale & Happywhale & 428 & 197 \\
short\_fin\_pilot\_whale+fin\_dorsal & contributed& 2,137 & 611 \\
short\_finned\_pilot\_whale & Happywhale & 745 & 458 \\
snow\_leopard & contributed& 4,715 & 241 \\
spinner\_dolphin & Happywhale & 2,254 & 517 \\
spotted\_dolphin & Happywhale & 490 & 280 \\
turtle\_green & contributed& 5,640 & 109 \\
turtle\_green+head & contributed& 6,282 & 109 \\
turtle\_hawksbill & contributed& 5,906 & 109 \\
turtle\_hawksbill+head & contributed& 7,625 & 112 \\
turtle\_loggerhead\_ext & SeaTurtleID & 7,405 & 438 \\
turtle\_loggerhead\_ext+head & SeaTurtleID & 7,166 & 438 \\
whale\_fin+fin\_dorsal & contributed& 2,998 & 942 \\
whale\_grey & contributed& 8,294 & 2,343 \\
whale\_humpback+fin\_dorsal & contributed& 9,251 & 3,924 \\
whale\_humpback+fluke & contributed& 67,06 & 4,767 \\
whale\_orca & contributed& 4,045 & 1,208 \\
whale\_orca+fin\_dorsal & contributed& 5,781 & 1,494 \\
whale\_sperm+fluke & contributed& 27,001 & 667 \\
whaleshark & contributed& 6,825 & 539 \\
white\_shark+fin\_dorsal & contributed& 1,771 & 484 \\
white\_sided\_dolphin & Happywhale & 229 & 175 \\
wilddog & contributed& 9,522 & 2,028 \\
zebra\_grevys & contributed& 12,267 & 626 \\

\bottomrule
\end{tabular}}
\end{table}

\subsection{Preprocessing and Labeling}

Images were preprocessed to detect animals of interest, locate them with an enclosing axis-aligned bounding box, and label their species. Low quality images were filtered out, although in many cases expert curators completed this before contributing images.  The result is a set of 0 or more ``annnotations'' per image, each a sighting of an individual animal.  
Each annotation also has its predominant viewpoint assigned --- left, right, front or back. As mentioned above, each  annotation has been assigned an individual ID through a curation process that involved expert manual decision making, sometimes using the guidance of an earlier algorithm to rank-order and suggest potential matching annotations.  These ID-ed annotations form the ground truth dataset from which the work in this paper starts.

\subsection{Viewpoint}

Viewpoint plays an important role in our experiments. For some species, such as belugas, there is only one (aerial / top) viewpoint. For many others, identifying information from different sides of the body --- such as the spots on a cheetah or a giraffe --- are so different that matching left and right views is impractical. For these, annotations from different viewpoints are considered different individuals in our dataset. For still other species, where the identifying information is predominantly on the outline, such as for the shape of the dorsal fin, opposite side views can match and therefore are not considered different animals for the purposes of our study.

\subsection{Data Splitting}

The dataset is divided into training and test sets without a separate holdout set. To assess model performance on both closed-set and open-set scenarios, we carefully designed the test set composition. Specifically, 50\% of the individuals in the test set are also present in the training set, but with different annotations (i.e., different images). The remaining 50\% of individuals in the test set were not part of the training set at all. This approach allows us to evaluate the model's ability to generalize to both known and unseen individuals.

To ensure the test set is balanced and to prevent the metrics from being skewed by individuals with a high number of annotations, we implemented several controls. The number of samples per individual was balanced across the test set, and the annotations in the test set were subsampled to include a maximum of 10 samples per individual.

The train/test split was performed with careful consideration of dataset size and individual representation. The selection process is a follows:
\begin{enumerate}
    \item For each individual with k annotations, we first determine whether the individual will be placed in the training or test set. A smaller training percentage is favored for smaller datasets to ensure a sufficient number of individuals in the test set.
    \item If an individual is selected for the training set, a subset of their annotations is reserved for the test set.
    \item If an individual is not selected for the training set, all of their annotations are used for the test set.

\end{enumerate}

Post-Split Filtering

\begin{enumerate}
    \item Training Set: Individuals with fewer than 3 annotations are excluded. There is no upper limit on the number of annotations per individual.
    \item Test Set: Individuals with fewer than 2 annotations in the test set are excluded. For individuals with more than 10 annotations, random 10 annotations are picked to remain in the test set.  Finally, only one annotation per encounter is allowed in the test set to avoid inflating performance through matching of near identical images. 
\end{enumerate}
A separate validation set was not utilized in these experiments due to the limited data available for a significant number of species. The small sample sizes made it impractical to generate train, validation, and test splits of reasonable size, particularly for robust evaluation. Instead, model hyperparameters were optimized using data splits generated from random seeds that differed from those used in the final experiments, ensuring that the test set remained unseen during tuning.

\subsection{Evaluation}

In evaluating the re-identification system, we employ a \textit{one-vs-all} evaluation scheme, which avoids the pitfalls of traditional approaches that use separate query and reference sets. While typical methods reserve a reference set of labeled individuals and treat the query set as unknown, there is often insufficient data to create independent query and reference sets that are entirely separate from the training data. Previous works \cite{megadescriptor, Adam_2024_WACV, wildfusion, Cheeseman2022-vd} use the training set as the reference set and the test set as the query set. This introduces a \textit{soft data leak}, as the model has been optimized on the reference set, even though it has not directly seen the query data. To mitigate this issue, we use the test set as both the query and reference set, while ensuring that annotations are never matched to themselves. The evaluation process is as follows:

For each annotation $A$ in the test set: 
\begin{enumerate} \item Select annotation $A$ as the query. \item All other annotations in the test set, excluding $A$, are used as the reference set. \item Compute accuracy at top-$k$ ranks for the query $A$. \item Repeat the process for all annotations in the test set. \end{enumerate}

Metrics are averaged across all annotations in the test set to generate the final evaluation result.

This approach ensures that the model is evaluated on unseen data without introducing biases from the training set. The final scores, averaged across all test annotations, provide robust performance metrics, yielding a more accurate evaluation of the model’s ability to generalize to unseen data.

\section{Model and Inference}

Our model model starts from an EfficientNetV2-M \cite{tan2021efficientnetv2} backbone pretrained on ImageNet1K \cite{imagenet15russakovsky}.  It is trained using a sub-center ($k=3$) ArcFace loss function \cite{deng2020subcenter} with dynamic margins \cite{ha2020googlelandmarkrecognition2020}. As will be discussed briefly in the experimental section below, unlike the MegaDescriptor \cite{megadescriptor} work, we find that using vision transformers did not improve the results. Each annotation's bounding box is used to crop the image and this crop is then scaled to a standard size prior to input. The production version size is 440x440 pixels, although for efficiency reasons the comparison experiments all use a smaller 256x256 resolution. Details on training and parameter settings are found in the appendix.

At inference time, the classification layer is discarded and an embedding vector of dimension 2,048, the default for EfficientNetV2 \cite{tan2021efficientnetv2}, is extracted when the trained model is applied to an annotation.  A test database of the embedding vectors is formed for each species separately, and each test annotation's vector is matched against the other vectors in the database and the top-k nearest are returned in order of increasing cosine distance, regardless of identity label.

\section{Experiments}

We present a range of experiments showing the importance of multispecies training for individual animal id, demonstrating the power of using this large diverse dataset for single species id, for adding new species, and for fine-tuning.  
\FloatBarrier
\begin{figure*}[!t]
    \centering
    \includegraphics[width=\linewidth]{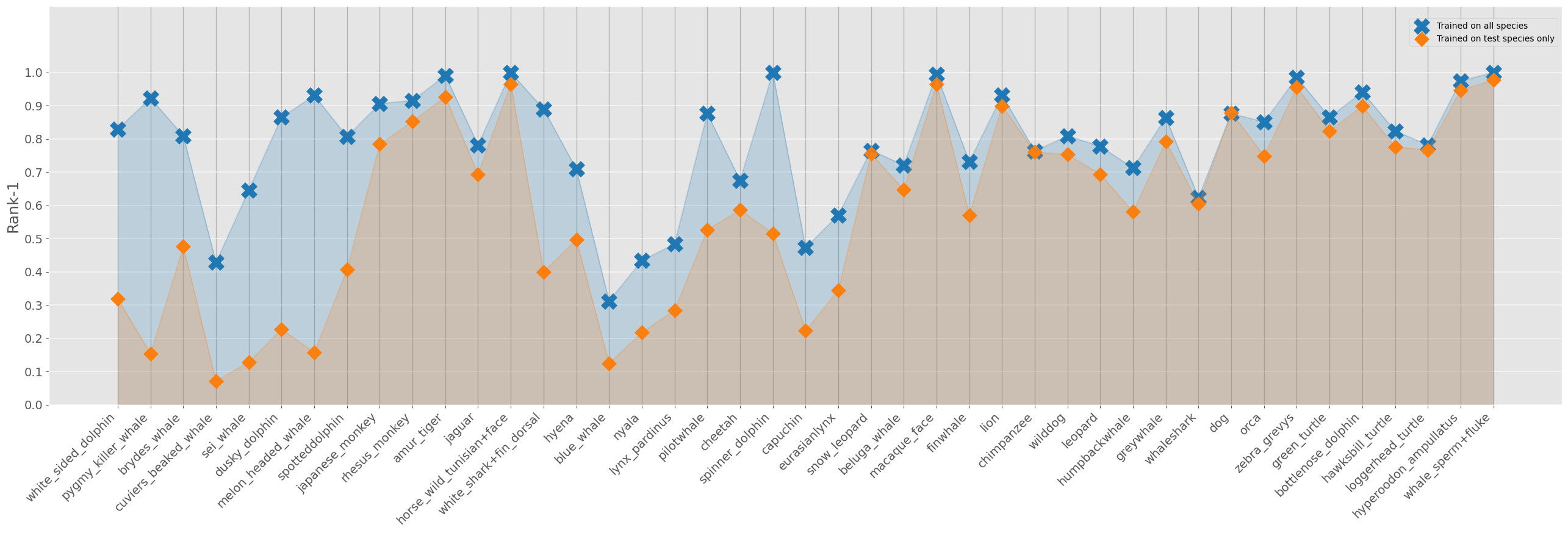}
    \caption{Comparison of top-1 performance of multi-species and single-species models. Species are ordered by increasing numbers of sightings.}
    \label{fig:multi-vs-single}
\end{figure*}

\begin{figure*}[!t]
    \centering
    \includegraphics[width=\linewidth]{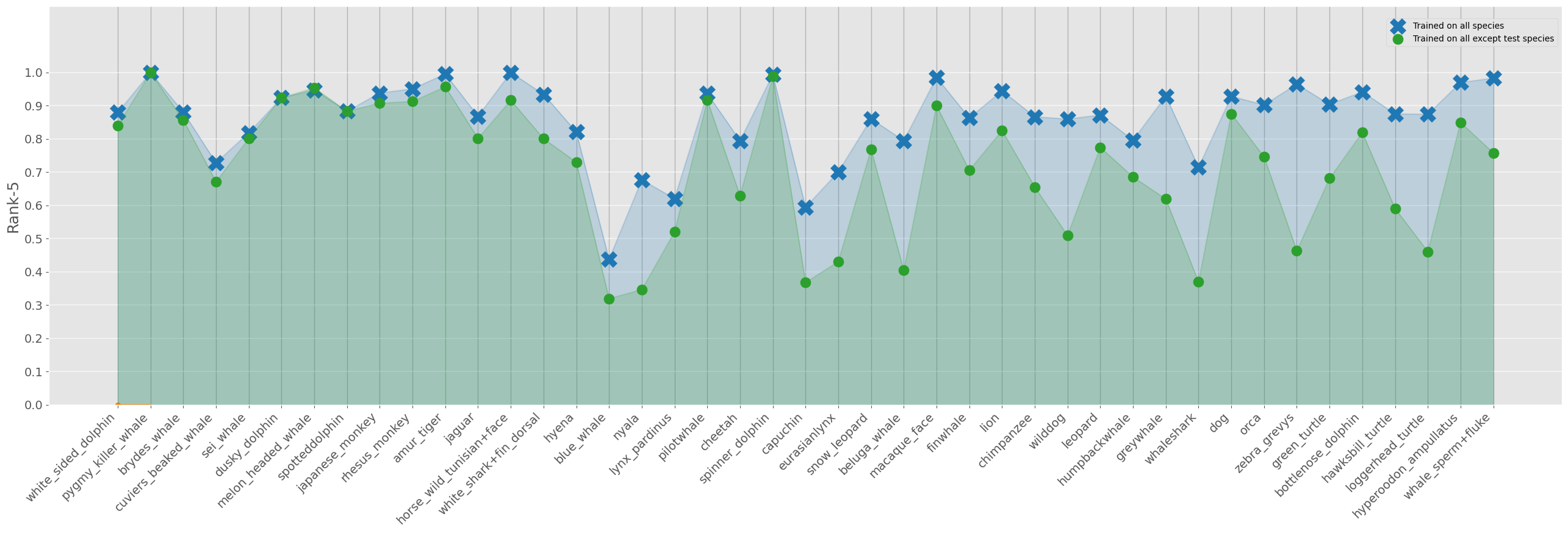}
    \caption{Top-5 performance comparison per species for model trained on the full dataset and model trained on all species except the test species}.
    \label{fig:loa-vs-ms}
\end{figure*}

\subsection{Multispecies vs.\ Individual Models}

Our largest experiment compares the performance of the full model trained on $N=49$ species against that of $N$ other models, each trained with the same architecture, loss function, and hyperparameters on each species separately. Top-1 ranking performances are shown in Figure~\ref{fig:multi-vs-single} and Table~\ref{table:selected_metrics}, with species ordered by decreasing number of annotations. These results show the multispecies model consistently outperforming single-species models, by an average of 12.5\% top-1 accuracy. The advantages of the multispecies model increase as the size of the population for any one species decreases (Figure~\ref{fig:multi-vs-single}). These results immediately give experimental justification for training and applying a single model across many species, an important practical consideration for managing multiple species at once. We are already seeing this in production use.

\subsection{Species Not Seen During Training}\label{loa}

One potential application of the multispecies model is to bootstrap identities for individuals from species with a paucity of labeled data. Our first study of this is the zero-shot case where no annotations from the species are seen during.  To study this, we train $N$ additional models, each with a single species left out, and measure the test performance on the left-out species. Figure~\ref{fig:loa-vs-ms} shows the results comparing the full model with the leave-one-out models. We show top-5 performance because that is more indicative of what's needed to guide human decision making when building up new species. (Top-1 results are provided in the supplementary material.) With species ordered by increasing numbers of annotations, the results show that for species with fewer annotations, the model uses aggregate information from training on many species to fill in missing information for matching. This effect is less prominent for species with larger datasets.

\begin{table}[htbp]
\caption{Comparison of metrics across various species for multi-species, leave-one-out and single species models}\label{table:selected_metrics}
\centering
\resizebox{\columnwidth}{!}{
\begin{tabular}{l l l l l }
\toprule
Source & Multi-species & Leave-one-out & Single species \\ \midrule
amur\_tiger & \textbf{99.1} & 92.3 & 92.7 \\
beluga\_whale & \textbf{72.1} & 28.0 & 64.8 \\
blue\_whale & \textbf{31.3} & 19.9 & 12.5 \\
bottlenose\_dolphin & \textbf{94.2} & 71.5 & 89.9 \\
brydes\_whale & \textbf{81.0} & 69.7 & 47.6 \\
capuchin & 47.3 & 22.3 & 22.3 \\
cheetah & \textbf{67.6} & 49.5 & 58.7 \\
chimpanzee & 76.4 & 45.1 & 76.1 \\
cuviers\_beaked\_whale & 42.9 & 37.3 & 7.1 \\
dog & 87.7 & 75.1 & \textbf{88.0} \\
dusky\_dolphin & \textbf{86.5} & 84.8 & 22.8 \\
eurasianlynx & \textbf{57.1} & 25.2 & 34.6 \\
finwhale & \textbf{73.2} & 49.2 & 57.1 \\
giraffe & \textbf{100.0} & 0.0 & 93.6 \\
green\_turtle & \textbf{86.5} & 48.3 & 82.4 \\
greywhale & \textbf{86.5} & 44.4 & 79.3 \\
hawksbill\_turtle & \textbf{82.5} & 38.1 & 77.6 \\
horse\_wild\_tunisian+face & \textbf{100.0} & 86.8 & 96.6 \\
humpbackwhale & \textbf{71.3} & 51.6 & 58.1 \\
hyena & \textbf{71.0} & 58.9 & 49.7 \\
hyperoodon\_ampullatus & \textbf{97.5} & 72.9 & 94.7 \\
jaguar & \textbf{78.2} & 67.8 & 69.4 \\
japanese\_monkey & 90.7 & 84.6 & 78.5 \\
leopard & \textbf{77.8} & 61.3 & 69.4 \\
lion & \textbf{93.2} & 75.2 & 90.0 \\
loggerhead\_turtle & 78.2 & 32.7 & 76.7 \\
lynx\_pardinus & \textbf{48.4} & 32.3 & 28.5 \\
macaque\_face & \textbf{99.5} & 71.8 & 96.5 \\
melon\_headed\_whale & \textbf{93.1} & 92.1 & 15.8 \\
nyala & \textbf{43.4} & 15.6 & 21.8 \\
orca & \textbf{85.1} & 59.5 & 74.9 \\
pilotwhale & \textbf{87.7} & 81.3 & 52.7 \\
pygmy\_killer\_whale & \textbf{92.3} & 87.6 & 15.4 \\
rhesus\_monkey & \textbf{91.6} & 74.5 & 85.3 \\
seal & \textbf{67.6} & 63.9 & 64.1 \\
sei\_whale & \textbf{64.7} & 62.5 & 12.9 \\
snow\_leopard & 76.6 & 65.2 & 75.5 \\
spinner\_dolphin & \textbf{100.0} & 95.2 & 51.6 \\
spotteddolphin & \textbf{80.7} & 74.8 & 40.8 \\
whale\_sperm+fluke & \textbf{100.0} & 62.0 & 97.7 \\
whaleshark & \textbf{62.4} & 24.2 & 60.5 \\
white\_shark+fin\_dorsal & \textbf{89.0} & 64.9 & 40.1 \\
white\_sided\_dolphin & \textbf{83.0} & 80.0 & 32.0 \\
wilddog & \textbf{80.9} & 40.2 & 75.3 \\
zebra\_grevys & \textbf{98.5} & 33.8 & 95.7 \\
\bottomrule
\end{tabular}}
\end{table}

\subsection{Our Model vs.\ MegaDescriptor}

\begin{table}[h]
\caption{Comparison of 0-shot rank-1 performace of our model to MegaDescriptor}\label{table:vs_megadescriptor}
\centering
\resizebox{\columnwidth}{!}{
\begin{tabular}{l l l }
\toprule
Source & MegaDescriptor-L-384 & Our model \\ \midrule
blue\_whale & 19.2 & \textbf{19.9} \\
bottlenose\_dolphin & 17.7 & \textbf{71.5} \\
brydes\_whale & 66.7 & \textbf{81.0} \\
capuchin & 14.2 & \textbf{22.3} \\
cheetah & 38.9 & \textbf{49.5} \\
cuviers\_beaked\_whale & 18.6 & \textbf{42.9} \\
dog & 57.0 & \textbf{75.1} \\
dusky\_dolphin & 28.5 & \textbf{84.8} \\
eurasianlynx & 15.7 & \textbf{25.2} \\
finwhale & 24.9 & \textbf{49.2} \\
green\_turtle & 22.3 & \textbf{48.3} \\
greywhale & 32.0 & \textbf{44.4} \\
hawksbill\_turtle & 25.0 & \textbf{38.1} \\
horse\_wild\_tunisian+face & 86.3 & \textbf{86.8} \\
humpbackwhale & 43.7 & \textbf{51.6} \\
hyperoodon\_ampullatus & 35.3 & \textbf{72.9} \\
jaguar & 41.1 & \textbf{67.8} \\
japanese\_monkey & 50.8 & \textbf{84.6} \\
lion & 69.6 & \textbf{75.2} \\
lynx\_pardinus & 18.6 & \textbf{32.3} \\
melon\_headed\_whale & 38.2 & \textbf{92.1} \\
orca & 46.8 & \textbf{59.5} \\
pilotwhale & 51.8 & \textbf{81.3} \\
pygmy\_killer\_whale & 61.5 & \textbf{92.3} \\
rhesus\_monkey & 47.3 & \textbf{74.5} \\
sei\_whale & 39.7 & \textbf{64.7} \\
snow\_leopard & 58.8 & \textbf{65.2} \\
spinner\_dolphin & 19.1 & \textbf{95.2} \\
spotteddolphin & 65.0 & \textbf{74.8} \\
whale\_sperm+fluke & 47.3 & \textbf{62.0} \\
white\_shark+fin\_dorsal & 11.2 & \textbf{64.9} \\
white\_sided\_dolphin & 72.0 & \textbf{80.0} \\
wilddog & 31.2 & \textbf{40.2} \\
\bottomrule
\end{tabular}}
\end{table}

The significance of the results on unseen species can be better appreciated by comparing them with the results from MegaDescriptor \cite{megadescriptor}. Both MegaDescriptor-L-384 and our model are evaluated at their native resolutions—384px and 256px, respectively — and neither has been trained on the tested species. Table~\ref{table:vs_megadescriptor} presents the rank-1 test results per species between MegaDescriptor and our model, trained without including the species in the training set (as detailed in Section \ref{loa}). 
The comparison is conducted on 33 species because these are the ones MegaDescriptor has not been trained on. In this zero-shot setting, our model outperforms MegaDescriptor on every species, with an average performance gain of 19.2\%. We attribute this advantage in part to the size and diversity of our dataset, as well as the fact that it has been extensively curated through ongoing, production-level review by species experts.

\subsection{Bootstrapping New Species}

As noted in the introduction, a major challenge in individual animal identification is bootstrapping new species from a limited number of annotated examples. In this subsection, we address this by selecting a representative set of species and retaining only a small fraction of the annotations in the training set. For each species separately, we then recombine this with the training set from all other species and train the model. By conducting this experiment at various data scales for the target species, we analyze the impact of more and more  limited datasets. The results from Table~\ref{table:bootstrapping} clearly show the advantages of including species, even in small amounts, in multi-species training. Moreover, joint training consistently achieves higher accuracy for the target species than the equivalent single-species model. This suggests that the shared representation learned from multi-species data enhances the model's generalization to new species, even when annotations are scarce.

\begin{table}[H]
\caption{Comparison of training the model on single species vs.\ joint training of target species with the full multi-species dataset.}
\centering
\label{table:bootstrapping}
\resizebox{\columnwidth}{!}{
\begin{tabular}{l c c c c}
\toprule
Source & \# of Individuals & \# of added annots & Species-only & Joint Training \\ \midrule
\multirow{6}{*}{Beluga} & 0 & 0 & --- & 27.9  \\
 & 25 & 557 & 32.5 & \textbf{46.7}  \\
 & 50 & 990 & 42.4 & \textbf{56.0} \\
 & 100 & 1,541 & 57.3 & \textbf{64.3} \\
 & 200 & 2,210 & 63.9 & \textbf{67.0} \\
\hline
\multirow{6}{*}{Hyena} & 0 & 0 & --- & 58.8 \\
 & 25 & 254 & 32.4 & \textbf{62.8} \\
 & 50 & 558 & 40.6 & \textbf{63.3} \\
 & 100 & 713 & 46.3 & \textbf{66.6} \\
 & 200 & 1,154 & 48.0 & \textbf{70.4} \\
\hline
\multirow{6}{*}{Loggerhead Turtle} & 0 &  & --- & 31.2 \\
 & 25 & 3,632 & 35.7 & \textbf{36.7} \\
 & 50 & 5,552 & 45.0 & \textbf{47.2} \\
 & 100 & 8,136 & 53.0 & \textbf{62.3} \\
 & 200 & 11,084 & 60.9 & \textbf{73.8} \\
\hline
\multirow{6}{*}{Gray Whale} & 0 & 0 & --- & 40.4 \\
 & 25 & 552 & 25.7 & \textbf{54.8} \\
 & 50 & 995 & 40.4 & \textbf{61.6} \\
 & 100 & 1,773 & 56.9 & \textbf{70.8} \\
 & 200 & 2,992 & 67.1 & \textbf{75.3} \\
\hline
\multirow{6}{*}{White Shark} & 0   & 0 & --- & 64.9 \\
                             & 25  & 331  & 21.1  & \textbf{77.6} \\
                             & 50  & 526  & 25.1  & \textbf{81.3} \\
                             & 100 & 864 & 31.0  & \textbf{85.4} \\
                             & 200 & 1,067  & 38.8  & \textbf{86.3} \\
\hline
\multirow{6}{*}{Lion} & 0 & 0 & --- & 74.5 \\
& 25 & 1,364 & 67.7 & \textbf{81.7} \\
& 50 & 2,180 & 74.4 & \textbf{84.7} \\
& 100 & 3,094 & 85.2 & \textbf{87.2} \\
& 200 & 3,846 & 89.2 & \textbf{90.1} \\
\bottomrule
\end{tabular}}
\end{table}

\begin{figure}[H]
    \centering
    \includegraphics[width=1\linewidth]{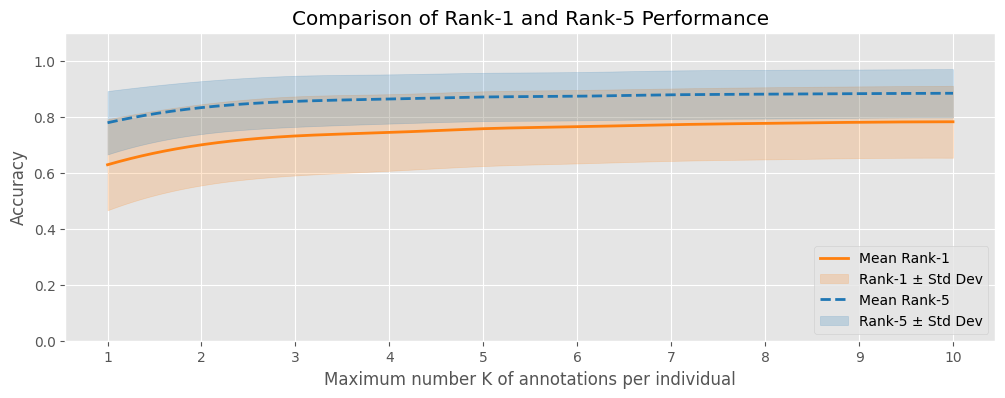}
    \caption{Mean top-1 accuracy and standard deviation across species as a function of the maximum number of annotations per individual in the database. }
    \label{fig:topk-filt}
\end{figure}

This ability to ``bootstrap'' new species can be further studied by showing the model's effectiveness in adding new individuals to a population. Here we are not interested in the ability to recognize that an annotation shows a new animal --- while an important problem, it is not our focus --- but rather how easily annotations are identified when there are $1, 2, 3, \ldots$ annotations in the database to correctly match a query annotation. We explore this for the same subset of species by showing average top-1 and top-5 rates for different numbers of potentially matching annotations, as shown in Figure~\ref{fig:topk-filt}. Matching performance is lower when there are less (k=1, k=2) annotations to match against in the database. However, the drop is not too significant, and performance quickly asymptotes, indicating strong potential for using the model to bootstrap new species since it does not require many sightings of an individual to reliably identify it.

\vspace{-2em} 

\vspace{-2em} 

\vspace{1em}

\subsection{Fine-Tuning on New Species}

\begin{table}[H]
\caption{Comparison of training the model from scratch vs.\ fine-tuning the multi-species model on varying amounts of individuals in the training set}
\label{table:fine-tune}
\centering
\resizebox{\columnwidth}{!}{
\begin{tabular}{l c c c c}
\toprule
Source & \# of Individuals & \# total of annots & Species-only & Fine-tuning  \\ \midrule
\multirow{6}{*}{Beluga} & 0 & 0 & --- & 27.9  \\
 & 25 & 557 & 32.5 & \textbf{44.5}  \\
 & 50 & 990 & 42.4 & \textbf{49.9} \\
 & 100 & 1,541 & \textbf{57.3} & 54.2 \\
 & 200 & 2,210 & \textbf{63.9} & 56.5 \\
\hline
\multirow{6}{*}{Hyena} & 0 & 0 & --- & 58.8 \\
 & 25 & 254 & 32.4 & \textbf{56.3} \\
 & 50 & 558 & 40.6 & \textbf{61.9} \\
 & 100 & 713 & 46.3 & \textbf{62.5} \\
 & 200 & 1,154 & 48.0 & \textbf{65.4} \\
\hline
\multirow{6}{*}{Loggerhead Turtle} & 0 &  & --- & 31.2 \\
 & 25 & 3,632 & 35.7 & \textbf{41.3} \\
 & 50 & 5,552 & 45.0 & \textbf{48.6} \\
 & 100 & 8,136 & 53.0 & \textbf{60.5} \\
 & 200 & 11,084 & 60.9 & \textbf{70.6} \\
\hline
\multirow{6}{*}{Gray Whale} & 0 & 0 & --- & 40.4 \\
 & 25 & 552 & 25.7 & \textbf{53.5} \\
 & 50 & 995 & 40.4 & \textbf{56.6} \\
 & 100 & 1,773 & 56.9 & \textbf{62.7} \\
 & 200 & 2,992 & \textbf{67.1} & 66.6 \\
\hline
\multirow{6}{*}{White Shark} & 0   & 0 & --- & 64.9 \\
                             & 25  & 331  & 21.1  & \textbf{66.7} \\
                             & 50  & 526  & 25.1  & \textbf{72.3} \\
                             & 100 & 864 & 31.0  & \textbf{74.8} \\
                             & 200 & 1,067  & 38.8  & \textbf{75.9} \\
\hline
\multirow{6}{*}{Lion} & 0 & 0 & --- & 74.5 \\
& 25 & 1,364 & 67.7 & \textbf{79.1} \\
& 50 & 2,180 & 74.4 & \textbf{83.5} \\
& 100 & 3,094 & 85.2 & \textbf{85.8} \\
& 200 & 3,846 & \textbf{89.2} & 87.3 \\
\bottomrule
\end{tabular}
}
\end{table}

Next we explore the fine-tuning capabilities of our model.  The goal is to use our model without needing access to the full training set to quickly apply to new species.  To explore this in Table~\ref{table:fine-tune} we compare single species training to fine-tuning of our model --- starting from the version trained without the species --- for various numbers of annotations. This shows that fine-tuning is effective, especially for small datasets, with single-species training only showing better performance for larger numbers of individuals and annotations. Finally, comparing the right columns of Tables~\ref{table:bootstrapping} and~\ref{table:fine-tune} we see that incorporating the data directly into training the multispecies model produces consistently better results.

\subsection{Vision Transformers}

In our experiments, we evaluated the impact of using a vision transformer backbone, specifically SwinV2-Base \cite{liu2021swinv2}, and compared its performance to the standard EfficientNetV2-M backbone. While vision transformers have shown strong results in various vision tasks, EfficientNetV2 consistently outperformed SwinV2 in our setting, achieving an average performance improvement of 4.5\%. This suggests that, for our task, EfficientNetV2 provides better feature extraction and generalization capabilities compared to the transformer-based approach.

\begin{figure}[h]
    \centering
    \includegraphics[width=1\linewidth]{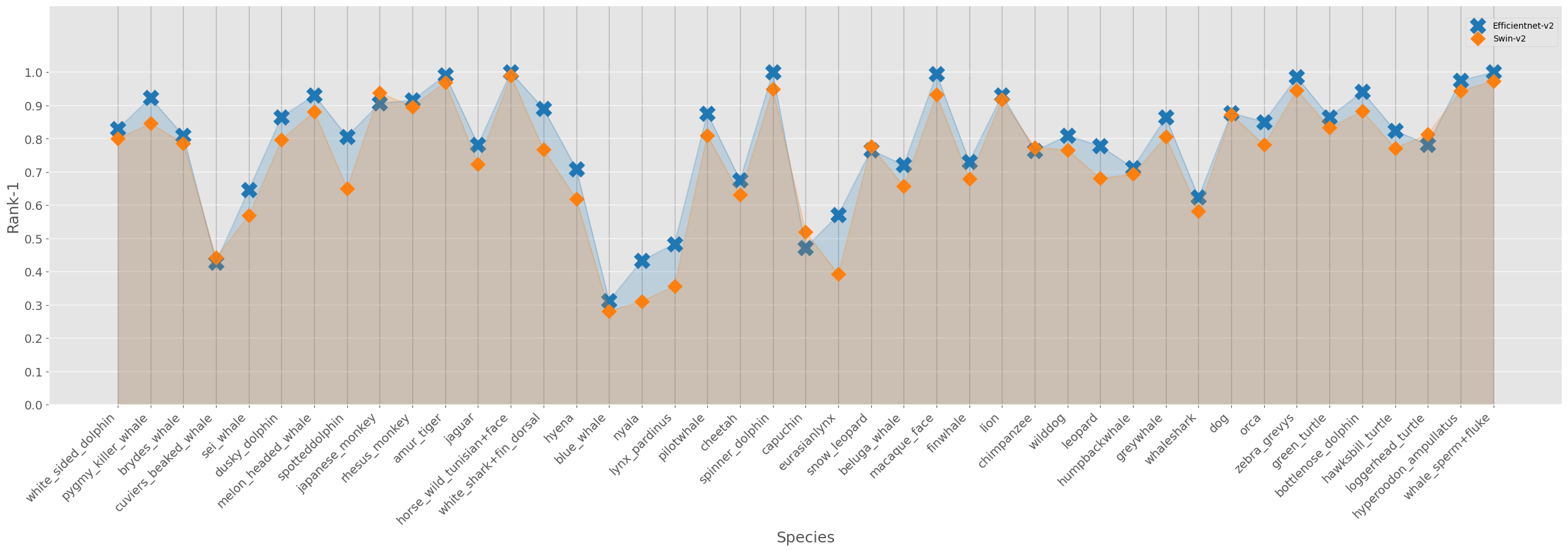}
    \caption{Comparison of Swin-V2 and Efficientnet-V2 backbones for MiewID training}
    \label{fig:enter-label}
\end{figure}

\section{Discussion and Conclusion}

We have presented and experimentally evaluated a multispecies individual animal identification model called MiewID.  Trained on 49 different species from 59 datasets, we showed that it consistently outperforms the same model trained on each species individually, with the greatest gains for species with small datasets.  Our model also demonstrated strong test performance on unseen (during training) species, uniformly and significantly better than the recent MegaDescriptor. Next, we showed that our model can be used as the starting point for fine-tuning, even with small dataset sizes. We also showed that incorporating these same datasets into multispecies training produced better results. This argues for frequent retraining as more and more species are added, which seems expensive but perhaps no more than training on each species individually, with a much simpler training and deployment process since just a single model is needed.

Our model is already in open-source production use on our Wildbook system.  More species are being added frequently and new models are periodically released on our \href{https://huggingface.co/conservationxlabs/miewid-msv2}{Huggingface page}.  The architecture, code, training procedure, and weights are freely available, and some of the datasets are as well. More should be distributed openly as the advantages of multispecies foundation models become clear.  Overall, we have achieved a significant advance in multispecies animal identification and intend to continue development as more and more well-curated datasets and advanced architectures and training methods become available.

\section{Acknowledgements}
The Gordon and Betty Moore Foundation directly supported our experiments. Precursor funding for cetacean-specific experiments was also received from the Bureau of Ocean Energy Management (BOEM) under awards 140M0121P0030 and 140M0121D0004. We would like to thank Dr. Ted Cheeseman for providing inspiration and community-sourced open cetacean data\cite{Cheeseman2022-vd} that benefited our work. We thank Paul Kalil and Maureen Reilly of T4C for coordinating the acquisition of several terrestrial carnivore datasets in Wildbook from a broad research community. Data used in this work was either publicly available or directly contributed to Wild Me via our hosted Wildbook platforms. Data contributors include: NOAA, Sarasota Dolphin Research Project, Botswana Predator Conservation Trust, Dr. Kasim Rafiq, Lion Guardians, Dr. Brad Brad Norman of ECOCEAN, Giraffe Conservation Foundation, WWF Spain, WWF Finland, Norwegian Orca Survey, Cascadia Research Collective, MarEcoTel, the Dominica Sperm Whale Project, African Parks, Ol Jogi Wildlife Conservancy, Ongava Research Centre, Endangered Wildlife Trust, Dalhousie University, Dr. Salvador Jorgensen, Olive Ridley Project, Dr. Eve Bohnett, Office français de la biodiversité, San Diego Zoo, and Cheetah Conservation Foundation.
C.\ Stewart's contributions were also partially supported by the US National Science Foundation under Award 2118240.

{\small
\bibliographystyle{ieee_fullname}
\bibliography{egbib}
}

\clearpage
\onecolumn
\appendix
\section{Appendix}
\subsection{Training}

Optimized with optuna.

Augmentation: rand color sharpening, clahe, shift (25\%), scale (20\%) and rotation (15 degree), color jitter

Learning rate schedule: linear warmup up in first 15 epoch (1.5e-5, 1.5e-3),  with exponential decay 0.8 per epoch

Batch size: 112

Image size: 256

Subcenter ArcFace with dynamic margins: k = 3, scale of 51.5, and margin is dynamic (started at 0.5), but calculated by class frequency.

Backbone is EfficientNetV2-M pre-trained on ImageNet-1K.

Final layer: GeM pooling + BatchNorm

Feature dimension: 2,152
\FloatBarrier

\subsection{Top-K Evaluation Results}
\FloatBarrier

\begin{figure}[H]
    \centering
    \includegraphics[width=0.98\linewidth]{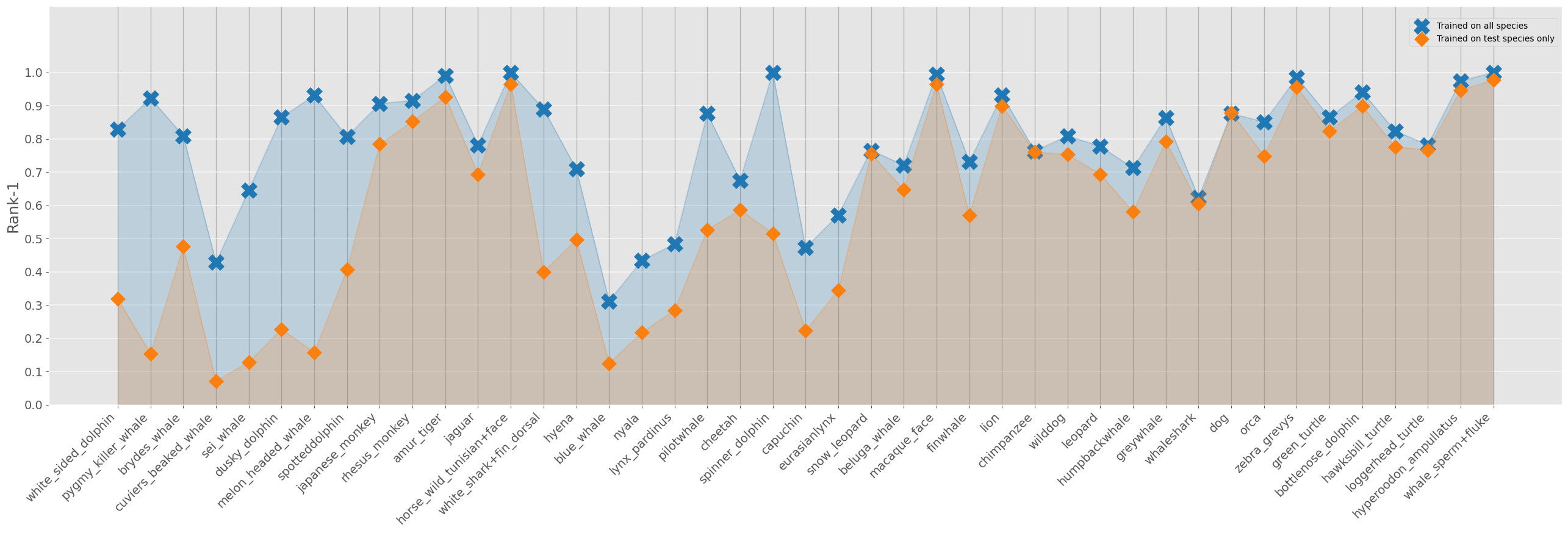}
    \caption{Comparison of top-1 performance of multi-species and single-species models. Species are ordered by increasing numbers of sightings.}
    \label{fig:multi-vs-single-r1}
\end{figure}

\FloatBarrier

\begin{figure}[H]
    \centering
    \includegraphics[width=0.98\linewidth]{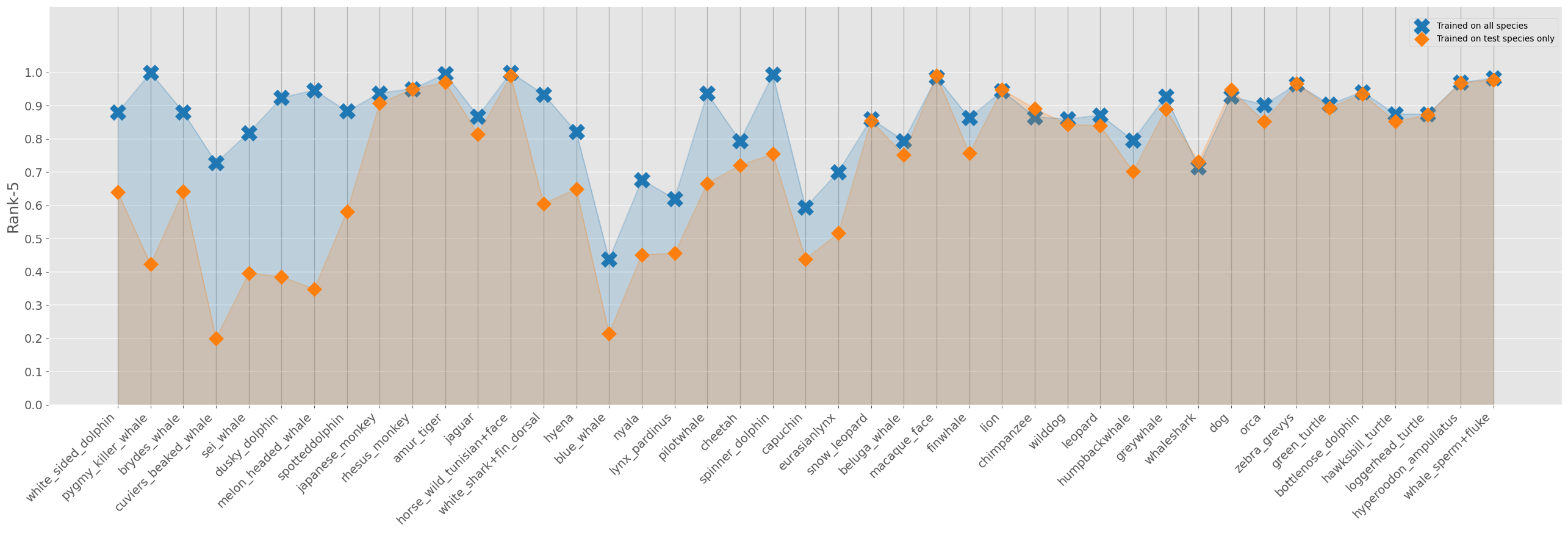}
    \caption{Comparison of top-5 performance of multi-species and single-species models. Species are ordered by increasing numbers of sightings.}
    \label{fig:multi-vs-single-r5}
\end{figure}
\begin{figure}[H]
    \centering
    \includegraphics[width=0.98\linewidth]{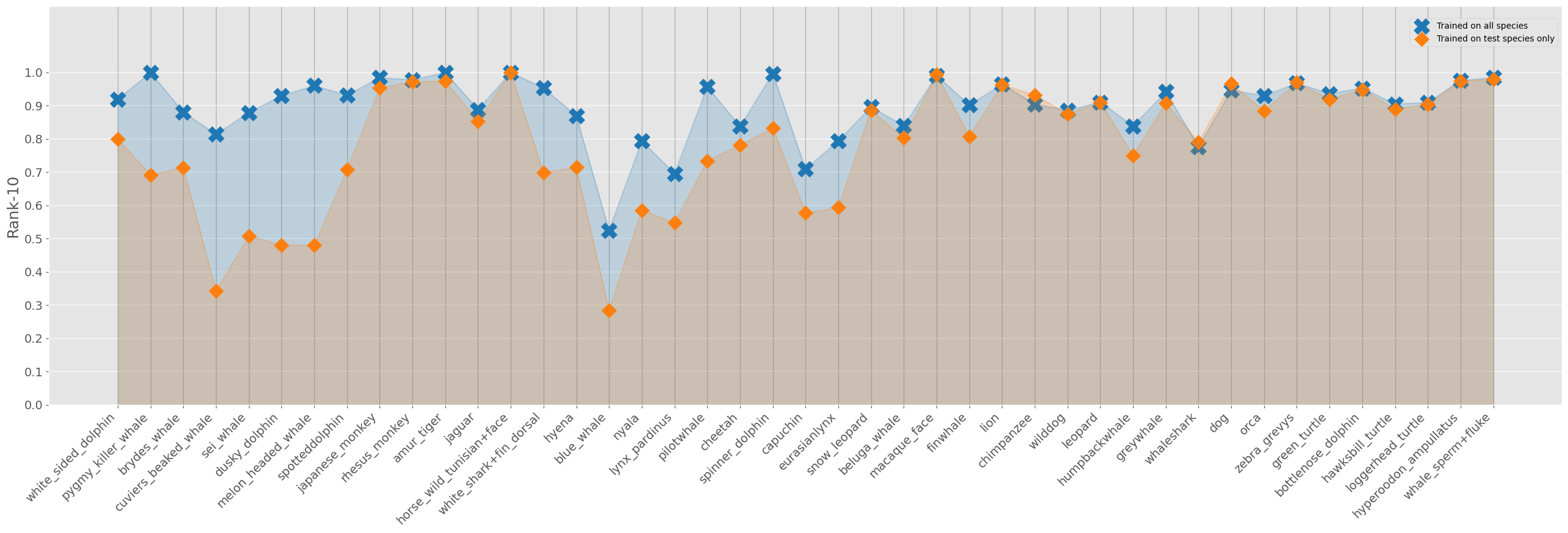}
    \caption{Comparison of top-10 performance of multi-species and single-species models. Species are ordered by increasing numbers of sightings.}
    \label{fig:multi-vs-single-r10}
\end{figure}

\begin{figure}[H]
    \centering
    \includegraphics[width=0.98\linewidth]{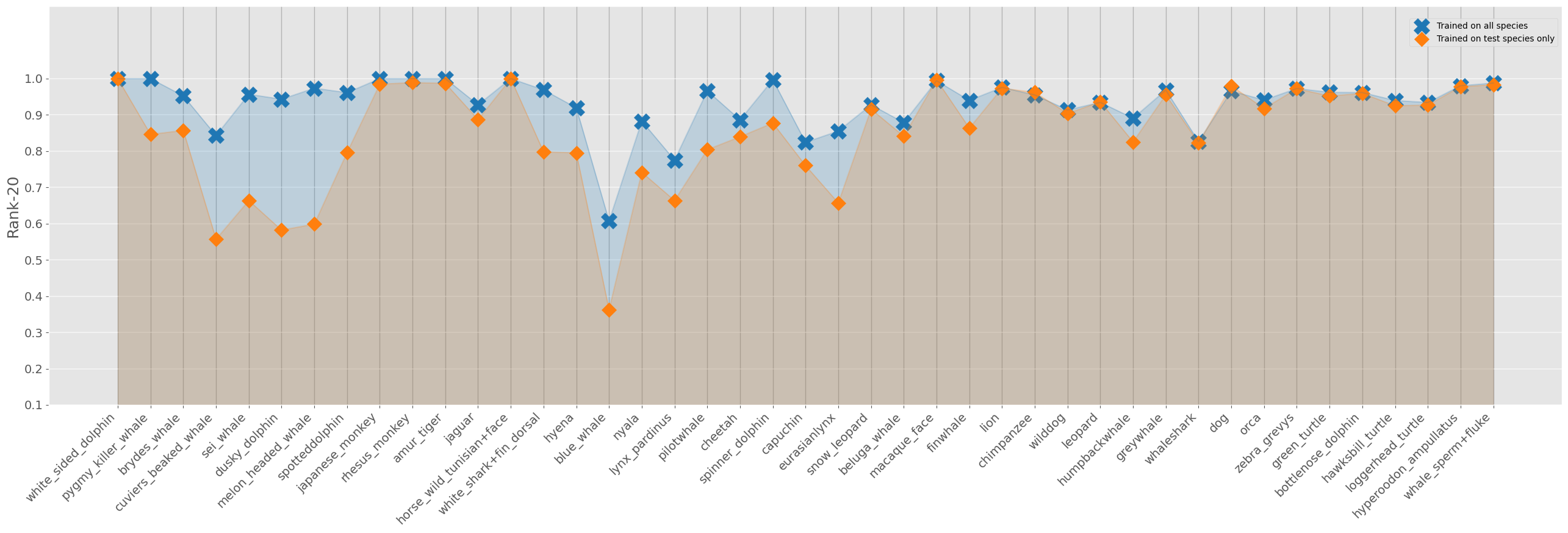}
    \caption{Comparison of top-20 performance of multi-species and single-species models. Species are ordered by increasing numbers of sightings.}
    \label{fig:multi-vs-single-r20}
\end{figure}

\begin{figure}[H]
    \centering
    \includegraphics[width=0.98\linewidth]{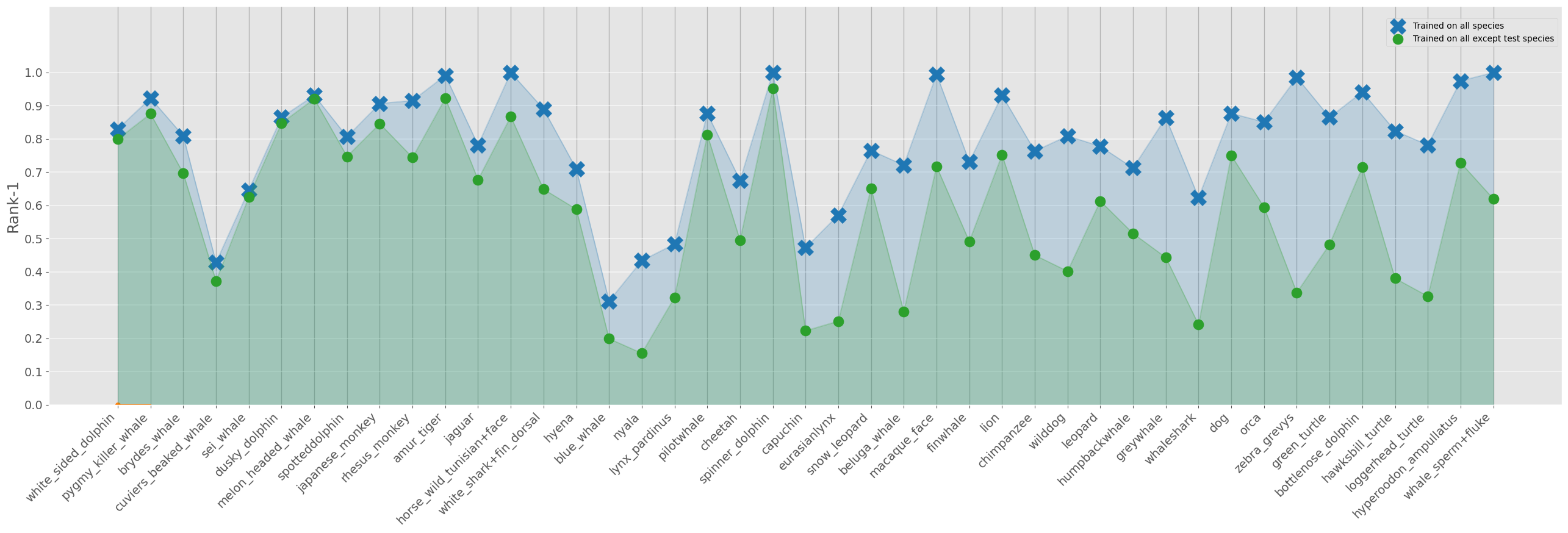}
    \caption{Top-1 performance comparison per species for model trained on the full dataset and model trained on all species \textbf{except the test species}.}
    \label{fig:loa-vs-ms-r1}
\end{figure}

\begin{figure}[H]
    \centering
    \includegraphics[width=0.98\linewidth]{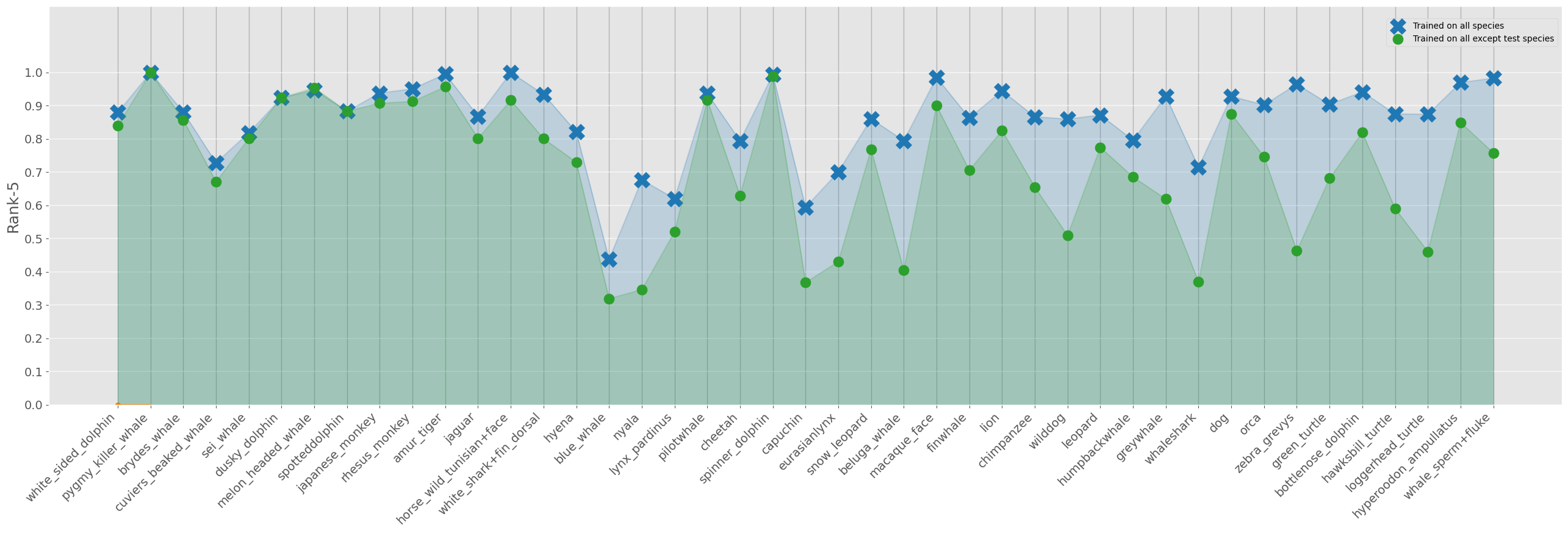}
    \caption{Top-5 performance comparison per species for model trained on the full dataset and model trained on all species \textbf{except the test species}.}
    \label{fig:loa-vs-ms-r5}
\end{figure}

\begin{figure}[H]
    \centering
    \includegraphics[width=0.98\linewidth]{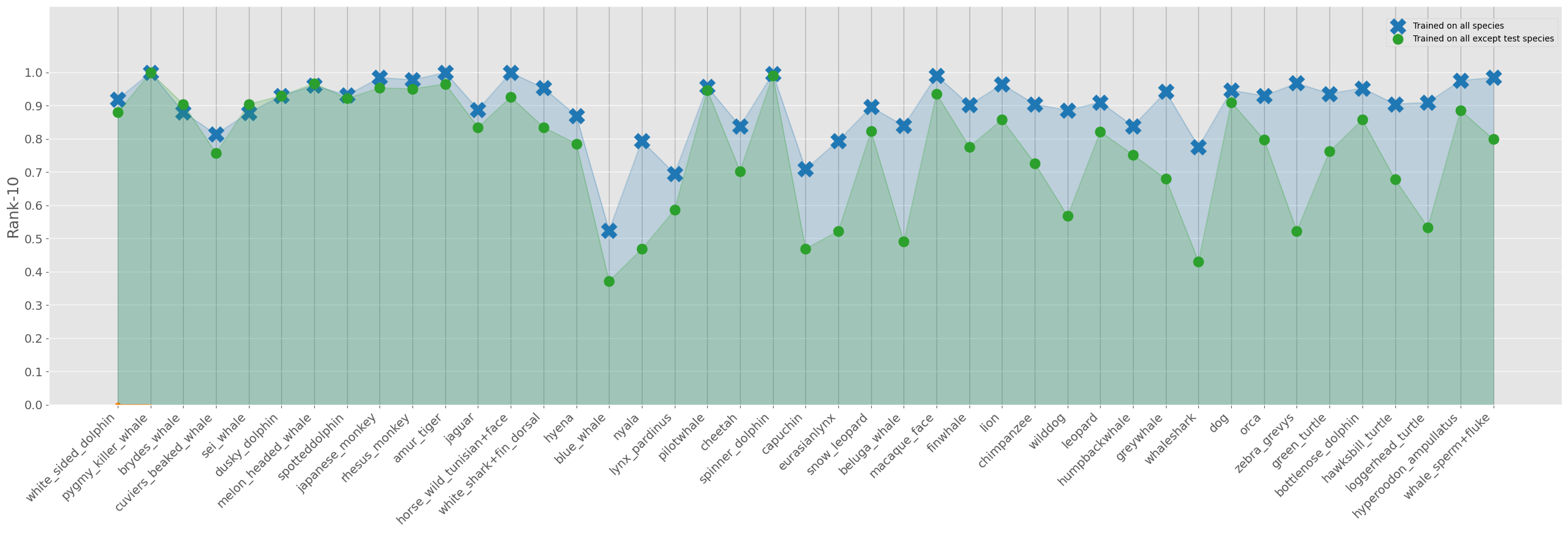}
    \caption{Top-10 performance comparison per species for model trained on the full dataset and model trained on all species \textbf{except the test species}.}
    \label{fig:loa-vs-ms-r10}
\end{figure}

\begin{figure}[H]
    \centering
    \includegraphics[width=0.98\linewidth]{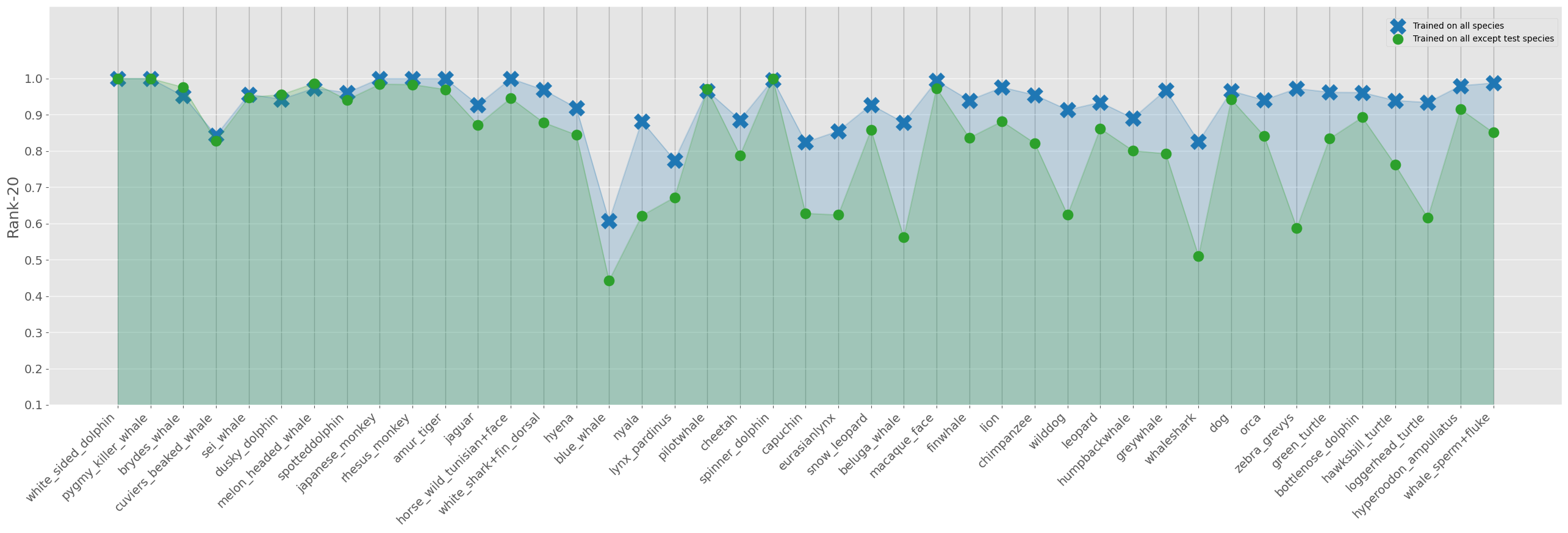}
    \caption{Top-20 performance comparison per species for model trained on the full dataset and model trained on all species \textbf{except the test species}.}
    \label{fig:loa-vs-ms-r20}
\end{figure}
\FloatBarrier

\end{document}